\documentclass{ecai2008}
\usepackage{times}
\usepackage{graphicx}
\usepackage{latexsym}

\def\ms#1{{\small $<$#1$>$}}
\def\me#1{{\small $<$/#1$>$}}
\def\manapher#1{\small ($\rightarrow$#1) }

\title{Symbolic Computing with Incremental Mind-maps to Manage and Mine Data Streams - Some Applications}

\author{Claudine Brucks \and Michael Hilker \and Christoph Schommer\\ \and Cynthia Wagner \and Ralph Weires \institute{University of Luxembourg, Campus Kirchberg, Dept. of Computer Science and Communication (CSC). MINE Research Group @ ILIAS Laboratory.  Address: 6, Rue Coudenhove Kalergi, 1359 Luxembourg, Luxembourg. Email: \{ \textit{name} . \textit{surname} \} @ uni.lu}
}

\begin{document}
\maketitle
\bibliographystyle{ecai2008}

\begin{abstract}
In our understanding, a mind-map is an adaptive engine that basically works incrementally on the fundament of existing transactional streams. Generally, mind-maps consist of symbolic cells that are connected with each other and that become either stronger or weaker depending on the transactional stream. Based on the underlying biologic principle, these symbolic cells and their connections as well may adaptively survive or die, forming different cell agglomerates of arbitrary size. In this work, we intend to prove mind-maps' eligibility following diverse application scenarios, for example being an underlying management system to represent normal and abnormal traffic behaviour in computer networks, supporting the detection of the user behaviour within search engines, or being a hidden communication layer for natural language interaction.
\end{abstract}

\section{ABOUT MIND-MAPS}\label{intro}

Transactional streams are to be understood as an endless flow of data that is lost once it is read. Furthermore, the data can be classified into categories, for example sentences or paragraphs inside a text document. These categories form the transaction, having items inside, for example words or paraphrases. An example for a transactional stream may be the reading of a book, where the text is read exactly once but lost if it is pronounced. The management, and moreover, the analysis of transactional streams is often problematic for several reasons. One of them is that data streams are potentially infinite or at least their end is not known until it is actually reached. Storing the whole data stream is therefore not an option, and the analysis cannot rely on traditional mining techniques that require the whole dataset to be available or that need random access or multiple passes over the data. Currently a lot of research focuses on the processing of such streams of different kind. Some of the typical techniques used with data streams are sliding windows, incremental approaches, or synopses of the data. Surveys of current methods and issues can be found in \cite{BAB02}, \cite{DOM01}, \cite{GOL03}, \cite{MUT03}.

With the discussion around \texttt{mind-maps}, we argue for its eligibility by demonstrating its applicability on a couple of algorithmic ideas. In our understanding, \texttt{mind-maps} are to be seen as adaptive and incremental knowledge structures, which live from depending on the occurrence of an input stream. A first approach in stream data analysis with mind-maps had been done in the processing of transactional streams with the creation of \texttt{mini-networks}. These base on transactional data (\cite{SCH04}), the mini-network consists of simple symbolic cells that share a weight value and that represent an individual item in a transaction. The symbolic cells are interconnected with other cells that occur in the transaction as well. In a subsequent step, the mini-network becomes integrated to the mind-map itself, where those cells become merged with those in the mind-map in case that they are identical (\texttt{merge}). Using the fundamental principle of adaptation and Hebbian Learning, the mind-map can be seen as a living engine as it is initially empty but grows over time. Since the states of connections and cells change over time the cells may die or revive as well. Focusing on the skeleton of the \texttt{mind-map}, a retrieve yields on delivering the strongest cell connections.

Although \texttt{mind-maps} often refer to such a structure to process fluid signals like text streams, mind-maps often claim to have its associative nature as the fundamental principle. This is true, however, we believe that a temporal character of such systems may be accepted as well to manage temporal stimulation that comes in. Secondly, mind-maps can be seen both from a verificative and an explorative perception: talking about mind-map software mostly refers to a top-down directed production of logically connected entities, for example work-flows or coherent cogitations (\cite{CHGM}, \cite{EPPL}, \cite{HAAS}), whereas the word of \texttt{explorative} more related to a learning or discovering process of such logical structures. In this respect, the following applications refers to the second class of mind-maps and present some algorithmic ideas for the temporal processing of text streams to map contextual information.  A simple implementation of mind-maps is the system \texttt{ANIMA}. It refers to a mind-map model of incremental and adaptive nature and allows to manage associations between symbolic cells while having a transaction stream as input. The aim of \texttt{ANIMA} is the efficient processing and management of transactions over time, to present related patterns inside a stream \cite{SCH04}. 

\section{APPLICATIONS}

\subsection{TARGETING NETWORK PROTECTION}\label{network}
Network-based anomaly detection \cite{HIL06} \cite{SUL96} refers to a system-based understanding of what the structure and the behaviour of network traffic is, and in this respect to identify abnormal situations. Here, the mind-map model \texttt{ANIMA-AR} is implemented to represent network traffic events while having network packets including header and content as one transaction, is one promising application. It fragments such packet transactions into meaningful symbolic cells within the mind-map and connects the packet cells. Additionally, connection values are established relating to the corresponding frequencies, respectively. Due to the architecture, we may rate the usual network traffic by a lower connection weight - although the frequency is high - but may rate an abnormal/unusual network behaviour by a higher value - although it appears more seldom. We identify the abnormal traffic as it is rated significantly and therefore tolerant against temporal connection updates between the symbolic cells. The following sequence of pictures gives an example on how the mind-map is used; it refers to the management of bad signatures as described in \cite{HIL06}, taking several insertion rules into account. First, the incoming signature ABC is considered and assigned to symbolic cells, being equal-weighted.

\begin{figure}[h]
   \centering
   \includegraphics[width=2.25cm]{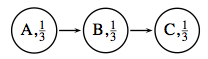} 
\end{figure}
   
Then, if a new signature CDEF is added to the mini-network, and the weights are adapted.  At each step, substrings are considered and evaluated as follows: given a sub-string, then if the sum of all activation states \dots

\begin{itemize}
    \item \dots is exactly 1, then a virus alert takes place. 
    \item \dots is exactly 0, then no virus alert takes place.
    \item \dots is between 0 and 1, an alert takes place with a probability value.
    \item \dots increases 1, then it is not considered.
\end{itemize}

For example, ABC forces a virus alert as it is infected for 100\%, whereas only H is unlikely to be infected. To continue the example, as \texttt{C} is already known, this value stays constantly.
 
\begin{figure}[h]
   \centering
   \includegraphics[width=4.5cm]{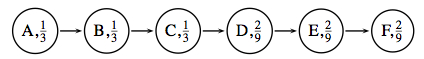} 
\end{figure}

The following photographs of the mind-map refers to the situation when the signatures \texttt{CDE} 

\begin{figure}[h]
   \begin{center}
   \includegraphics[width=4.5cm]{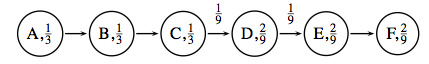} 
   \end{center}
   and \texttt{CDGH} are being inserted. With this, the probability values for each signatures is clearly available through the whole life-time of the mind-map. More information can be found in \cite{HIL06}.
   \begin{center}
   \includegraphics[width=4.5cm]{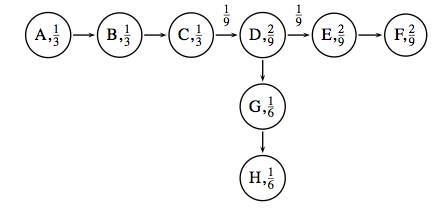} 
   \end{center}
\end{figure}

The mind-map model ANIMA-AR is implemented to detect well-known viruses. The signature of viruses is stored in a graph-like structure;  virus signatures are managed and stored, incoming packets - to identify intrusions - evaluated. The scanning speed and the required storage space outperforms current approaches and emerges out of the compression of the signature database. \texttt{ANIMA-AR} is theoretically analysed showing that viruses and similarities are detected. Simulations substantiate the theoretical analysis and show the low false-positive rate tolerating the normal system. In addition, ANIMA-AR is able to automatically detect similar viruses as small mutations or new variants.

\subsection{MANAGING IMPLICIT FEEDBACK}\label{retrieval}

The consideration of implicit feedback in the field of information retrieval and the automatic collection of information about the user's behaviour \cite{LPT99} \cite{WEI07} is an application of interest. Without an explicitly request of information, we intend to gain some information about what really interesting is. The aim is to use the mind-map as an adaptive storage for such kind of information and consequently, for the enhancement of user-based research requests. We therefore monitor the user's behaviour in interaction with a search engine, keeping an eye on queries and their results, links that the user follows, and diverse time-related information, for example how long he or she stays on web sites. Search sessions like this are considered as single transactions for the mind-map. And, building up such a network provides information about typical queries, results, and measures, especially regarding the relevance of search results, namely towards an enhancement of further queries. 

\begin{figure}[htbp]
   \centering
   \includegraphics[width=7cm]{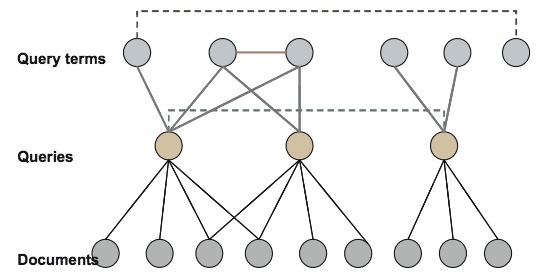} 
   \caption{Architecture of the mind-map \cite{WEI07}}
   \label{fig:sc10}
\end{figure}

An example might be a giving of additional hints or altering the results themselves. One concrete step is a re-ranking of search results, which may help to place those results further on the top of the list that are probably of greater importance according to the given query. Following this, the strengthens of the mind-map are mainly due to the ability to cope with transactional streaming data. We concern with information about search sessions of users, which can easily be broken down into transactions. Moreover, the mind-map is able to store only the most important aspects of the information without the need of storing all feedback data. This helps to keep the network at a reasonable size. Furthermore, the dynamic nature of this mind-map fits quite well to the purpose of this approach: if there is a change of the user's feedback over time, then this trend will be reflected in the mind-map as well. Figure \ref{fig:sc10} shows an architectural snapshot of the mind-map, where we use three different types of cells: 

\begin{itemize}
    \item The query terms are single terms that are observed in user queries.
    \item Queries that form the transactional input to the mind-map.
    \item The resulting list of documents that have been provided by the underlying search engine for one or more queries.
\end{itemize}

There may exist different connections between these nodes, which might be weighted indicating the strength of the relationship. These are for example connections that indicate relationships between different query terms, connections between queries and query terms, and connections between queries and documents. More information can be found in \cite{WEI07} \cite{WEI08}.

\subsection{TARGETING DBLP}

Bibliographical databases such as \texttt{Citeseer}, \texttt{Google Scholar}, and \texttt{DBLP} serve as a bibliographic source with lots of information concerning a publication. This compounds the names of the authors, the publication title, the conference, and many other attributes. A bibliography database is accessible online, where all entries share an electronic index to articles, journals, magazines, etc. containing citations and abstracts. Understanding the bibliographic database as a digital collection that is intelligently managed and that supports a search for and the retrieve of bibliographic information to defined queries is a convenient procedure in demanding information right in time. Regularly, the retrieve bases on a collection of queries that consists of keywords in the publication title or the keyword list. For example with \texttt{DBLP}, the querying using a keyword \texttt{plagiarism} leads to an answer set of almost 70 articles, and a search refinement with \texttt{detection} and \texttt{pattern} to 34 and 2 bibliographic entries, respectively. Accordingly, the two referenced publications pledge close; and the names of the authors overlap:

{\small
\begin{verbatim}
1 NamOh Kang, Sang-Yong Han: Document Copy
Detection System Based on Plagiarism Patterns.
CICLing 2006: 571-574.
2 NamOh Kang, Alexander F. Gelbukh, Sang-Yong
Han: PPChecker: Plagiarism Pattern Checker in
Document Copy Detection. TSD  2006: 661-667.
\end{verbatim}
}

With this, we reference to a graph structure representing the association of the three authors \texttt{Han}, \texttt{Kang} and \texttt{Gelbukh}. This is similar to \cite{KEM01} who introduces mini-networks. The double edges signalise a double connection as the single (and parallel) edge between \texttt{Han} to \texttt{Gelbukh} and \texttt{Kang} to \texttt{Gelbukh} refers to a one-way association. Whereas the meaning of a double connection is unique while having two publications, the double entries to \texttt{Gelbukh} seem to be ambiguous: on the one side, it refers to one common publication with both individually, on the other to two single publications with one of the other authors. However, the graph is node-oriented in a way that it simply represents the situation as it is: \texttt{Gelbukh} has one common publication with both of them, and here, it plays no role if this is a common one or not.

For static databases, the discovery of associative patterns has been an area of extensive research, and multiple approaches and solutions to the static problem have been presented in the past. A major problem in these approaches is the combinatorial explosion of the search space and the research has therefore mainly focused on reducing this space. Since mind-maps are targeted to data streams, it can not possibly make use of the methods developed for static databases, since these algorithms require multiple passes over the data to calculate frequencies of associated items. This is indeed not acceptable when dealing with data streams, because data streams are potentially infinite, or at the very least their end cannot be foreseen. In this respect, the idea of searching for temporal patterns in a bibliographic database like \texttt{DBLP} while taking the time as the core medium leads to a couple of interesting questions, for example

\begin{itemize}
   \item In general, may we discover scientific communities? While observing the visualisation of associative relationships between authors, we might ask if such dependencies generally form a community, and secondly, how strong these communities may be. Furthermore, if dependencies of of communities exist, are these temporal or visiting, recurring, or constant as it has been mentioned in section \ref{intro}?
   \item Do there exist diverse trends in publishing? For example, the occurrence of a common publication may be the initiator for a fruitful collaboration (which is proven by following publications on the same or a different research topic)
\end{itemize}

\begin{figure}[htbp]
   \centering
   \includegraphics[width=7cm]{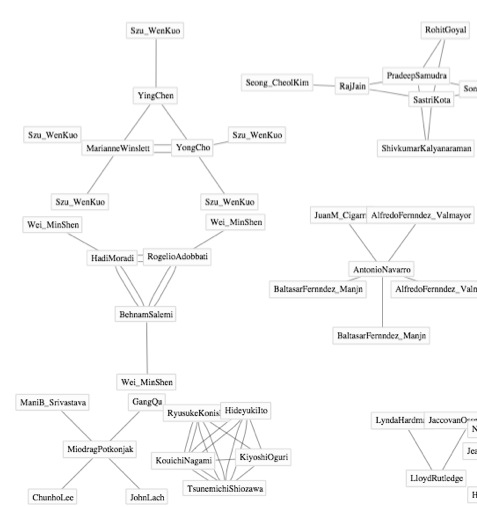} 
   \caption{Temporary mind-map (DBLP, year of 1993), consisting of associated author nodes.}
   \label{fig:dblp}
\end{figure}

With this, we may perform mind-mapping over a period of time, for example moving a corresponding window temporarily over time. 

\subsection{SEMANTIC NET-LEARNING}\label{wiwy}
The mind-map model \texttt{WYWI} stands for a simple communication paradigm that focus on natural language communication. The model uses a mind-map to manage words and their relationship to others associatively. A sentence is read incrementally and threatened as a transaction with \texttt{concepts} and \texttt{roles}. Currently, only adjectives, nouns, and verbs are considered as worth, they are extracted and put into the mind-map as a semantic structure. Adjectives are considered as sub-concepts of nouns. For example, 

{\small
\begin{verbatim}
#S(CONCEPT :NAME MAN :CAT N :FATHER (ROOT)
  :CHILDREN (YOUNG) :ROLES (READ) :ACT 0.9577)
#S(ROLE :NAME READ :CAT V :CONNECTION ((MAN
  BOOK 0.9577)) :ACT 0.9577)
#S(ROLE :NAME SEE :CAT V :CONNECTION ((LION
  PETER 0.8253)) :ACT 0.9014)
\end{verbatim}
}

Here, the word \texttt{READ} acts as relationship (\texttt{role}) connects both \texttt{MAN} and \texttt{BOOK}, sharing a connection and activation value of 0.9577, respectively. Other roles exist, but are unrelated. Each time-step, the activation values are decreased unless the role or the associated concepts remain inactive. As for the \texttt{ANIMA} mind-map model, the underlying net structure is refreshed after each transaction; it adapts to the given stimulation. However, \texttt{WYWI} does not use any extern source (for example thesaurus) but actively interacts with the user in case a word is unknown. This is to understand the morphologic basis or the lexical category. \texttt{WYWI} also establishes a semantic representation through micro-features: here, the user must rate a word semantically against given (predefined) entities. This produces a k-dimensional space, having k as the number of independent micro-features, for example

{\small
\begin{verbatim}
...
MAN is ((1.0 0.0 1.0 0.0 0.0 0.0 0.0 1.0 ...))
LION is ((0.0 0.0 0.0 0.0 0.7 0.0 0.4 ...))
...
\end{verbatim}
}

The communication in \texttt{WYWI} is performed as a conversation. For a sentence like \texttt{The yellow lion sees Peter}, the word \texttt{YELLOW} becomes a sub-concept of a word \texttt{LION}, whereas \texttt{LION} and \texttt{PETER} are connected by \texttt{SEE}. Additionally, \texttt{SEE} is becomes higher activated. Additionally, all concepts are represented in the k-space. As a consequence, transactional (dialogue) stream can be processed and queries be sent at each time.

{\small
\begin{verbatim}
...
WYWI: (ENTER THE NEXT WORD OR CLOSE BY '-')
USER: Peter
WYWI: COULD YOU PLEASE GIVE SOME EXPLANATION,
            LIKE THE MORPHOLOGIC BASIS?
USER: Peter
WYWI: AND WHAT IS THE LEXICAL BASIS OF peter?
USER: person
WYWI: I NEED SOME SEMANTIC INFORMATION ABOUT
      peter. HAVE A LOOK AND RATE peter FROM
      0 to 1:
      TALL?
USER: 0.9
WYWI: And with HEAVY?
...
\end{verbatim}
}


An extended version of the semantic net-learning from text streams is to define a temporal mind-map for a certain actor of a text: where the text stream is consolidated over time by a semantic graph structure. In this work, an important issue for building an author-related mind-map is the resolution of linguistic anaphors.

So far, the major idea is to process each sentence incrementally so that a general and pre-defined scheme structure of \texttt{subject - verb - object} is instantiated every time (\cite{LMG05}). At the moment, we only concern with nouns, verbs, and adjectives, where other linguistic categories are disregarded. In this respect, verbs act as a role between the concepts (noun, object). Each concept may have sub-concepts that correspond to attributes, for example the sentence \texttt{the old man likes the green juicy grassland} is translated into

\begin{figure}[!htbp]
\centering
\includegraphics[width=0.42\textwidth]{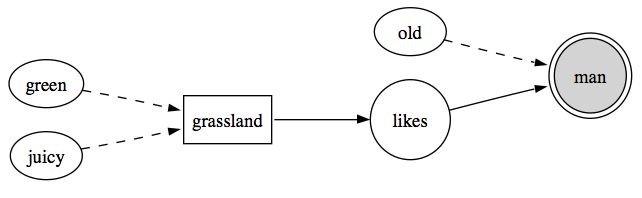}
\caption{A semantic graph having \texttt{man} as the centric concept. Attributes are adjunct by dashed lines, whereas roles are connected as a circle.}
\label{fig:Graph1}
\end{figure}

\begin{figure}[!htbp]
\centering
\includegraphics[width=0.42\textwidth]{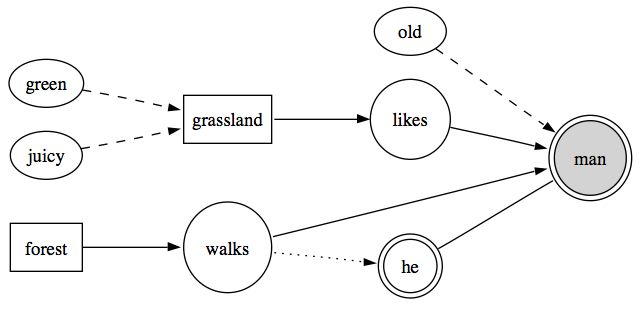}
\caption{The anaphor \texttt{he} is recognized as to be related to \texttt{man}, whereas the two sentence structures are merged by the concept \texttt{man}.}
\label{fig:Graph3}
\end{figure}

where \texttt{man} and \texttt{grassland} represent main concepts and \texttt{green}, \texttt{juicy}, and \texttt{old} sub-concepts. The process of relating identical concepts together is accomplished by a matching of identical words and a resolution of linguistic anaphors (\cite{LL94}, \cite{MIT98}). The following Figure \ref{fig:Graph3} shows the semantic structure after having read the sentence \texttt{He walks into the forest}.

So far, the incremental processing of texts can be stopped at any moment. The semantic structure is a mind-map with concepts and relationships of consistent states. Beside focusing on the elaboration of methods to realize pronominal anaphora and co-reference in text streams, we will assign weight values to concepts and roles in order to prove their importance for an actor. 

The idea of content zoning is to refer to a segmentation of a text document into semantic zones. As indicated in \cite{BW07} and moreover as firstly discussed in \cite{TE99} with \texttt{Argumentative Zoning}, the basic idea here is to structure texts on the basis on pre-defined categories. An example might be the following text, having the actors \texttt{Harry}, \texttt{Hedwig}, and \texttt{owl}:

{\small\it
``Harry got up off the floor, stretched, and moved across to his desk. Hedwig made no movement as she began to flick through newspapers, throwing them into the rubbish pile one by one. The owl was asleep or else faking; she was angry with Harry about the limited amount of time she was allowed out of her cage at the moment. As her neared the bottom of the pile of newspapers. Harry slowed down, searching for one particular issue that he knew had arrived shortly after he had returned to Privet Drive for the summer, he remembered that there had been a small mention on the front about the resignation of Charity Burbage, the Muggle Studies teacher at Hogwarts."} 

an then be zoned to 

{\small\it
\begin{itemize}
    \item Harry \ms{ACTOR=HARRY} got up off the floor, stretched, and moved across to his \manapher{his  $\Rightarrow$  HARRY} desk. \me{ACTOR=HARRY}
    \item Hedwig \ms{ACTOR=HEDWIG} made no movement as she \manapher{she = HEDWIG} began to flick through newspapers, throwing them \manapher{them = newspaperss} into the rubbish pile one by one.
    \item The owl \ms{ACTOR=OWL} was asleep or else faking; she \manapher{she = OWL} was angry with Harry \manapher{Harry = HARRY} about the limited amount of time she \manapher{she = OWL} was allowed out of her \manapher{her  $\Rightarrow$  OWL} cage at the moment. As her \manapher{her $\Rightarrow$ OWL} neared the bottom of the pile of newspapers, \me{ACTOR=OWL}. 
    \item Harry\ms{ACTOR=HARRY} slowed down, searching for one particular issue that he \manapher{he = HARRY} knew had arrived shortly after he \manapher{he = HARRY} had returned to Privet Drive for the summer, he \manapher{he = HARRY} remembered that there had been a small mention on the front about the resignation of Charity Burbage, the Muggle Studies teacher at Hogwarts. At last he \manapher{he = HARRY} found it.\me{ACTOR=HARRY}
   \end{itemize}
}

where linguistic anaphors are solved depending on the current actor, the gender, and/or between different candidates that have already occurred in the text. Knowing that the owl has the name Hedwig, and moreover, having a hierarchy on site, the zoning can be updated even more. An additional analysis can be performed to gain further information about the separate zones. Rather simple to extract information could be statistics about the size and layout of zones, but a more sophisticated analysis of their text content is possible. The latter can lead to an extraction of the semantic content and purpose of a zone. Such kind of information can be used for various purposes, such as comparing documents to each other (regarding their analysed zone structure and content) by simply refering to information about the zones as zone variables (\cite{BW07}).

We apply content zoning to text streams in order to establish a semantic mind-map, while using a sliding window of user-specified length; text is buffered in, immediately zoned and analysed. Intermediate statistical results are managed to produce a user specific summary on a given subject. The definition of zones is insufficient to an effective zoning as zones only indicate a position of an information in the text, but do less give information about the content. A solution might be the introduction of zone variables, which describe the content of the input stream and are the core parameters for the summary generation. In this respect, we concern with two categories of zones, namely \texttt{document independent} and \texttt{document dependent} zones, for example the position of the zone in the text stream, its length or the most occurring word (without stopwords). During the zoning process, nearly all sentences are attributed to zones. Useless sentences and sentences which cannot be attributed to a zone are skipped or can be regrouped a user-defined zone.  After different steps as anaphor resolution for example, the values of the variables are used for statistical evaluations to generate a summary on a user specific subject. During the process of buffering and processing the text streams, there is an option of real-time evaluation, so that changing values are immediately visible to the user. At each instant, the user has the possibility to call a summary on a previously defined subject, as for example an actor of a fairy tale. But, also less sophisticated results as the top most occuring words or collocations can be called at each moment.

\section{CONCLUSIONS}
A mind-map is an adaptive engine that basically works incrementally on the fundament of transactional streams. Following our model, mind-maps consist of symbolic neural cells that are connected with each other and that become either stronger or weaker depending on the transactional stream: based on the underlying biologic principle, these symbolic cells and their connections as well may adaptively survive or die, forming different cell agglomerates of arbitrary size.

With that, mind-maps may be applicable for the management of trust as well: every human shares an own attitude about others, for example \texttt{Person R} has an attitude to \texttt{Person S} and vice versa. Both are probably different from each other and different to \texttt{Person T}'s view to \texttt{Person R} or \texttt{Person S}. Furthermore, one might get the conclusion if \texttt{Person R} trusts \texttt{Person S} but not vice versa. With this approach, we may follow a proposition by \cite{MI05} who suggests a model on human conversations. The attitude of someone's mind is modelled as a self-organising mind-map. Every person has a model of his/her view of the world and models for other people whom he/she has interacted with. All views including a self view and views to other persons (others) will be modified through conversations between people. We therefore intend to introduce an engine to find regularities between human's objects and to model trust based on this. The creation of an artificial mind-map, where a textual data stream is read and represented in an associative dynamic network. Each incoming stream is decomposed to its items, for example a text stream may be decomposed to words.

\ack
This work has been done at the MINE research group at the Laboratory for Intelligent and Adaptive Systems across several research projects funded by the University of Luxembourg and the Ministry of Higher Education.

\end{document}